\documentclass[letterpaper, 10 pt, conference]{IEEEtran}
\IEEEoverridecommandlockouts

\usepackage{tikz}
\usepackage{graphics} %
\usepackage{epsfig} %
\usepackage{mathptmx} %
\usepackage{times} %
\usepackage{amsmath} %
\usepackage{amssymb}  %
\usepackage{subcaption}

\usepackage{pifont}
\usepackage[T1]{fontenc}
\usepackage{xspace}
\usepackage{xcolor}
\usepackage{multirow}
\usepackage{booktabs}
\usepackage{changepage}

\usepackage{newtxtext} %

\usepackage{hyperref}
\hypersetup{
    colorlinks=true,
    linkcolor=blue,
    filecolor=blue,      
    urlcolor=blue
}

\input{_macros}

\title{\LARGE \bf
ConfigBot: Adaptive Resource Allocation for Robot Applications in Dynamic Environments
}

\author{Rohit Dwivedula, Sadanand Modak, Aditya Akella, Joydeep Biswas, Daehyeok Kim, and Christopher J. Rossbach%
\\The University of Texas at Austin
}

\begin{document}

\maketitle

\begin{abstract}
The growing use of service robots in dynamic environments requires flexible management of on-board compute resources to optimize the performance of diverse tasks such as navigation, localization, and perception. %
Current robot deployments often rely on static OS configurations and system over-provisioning. However, they are suboptimal because they do not account for variations in resource usage. This results in poor system-wide behavior such as robot instability or inefficient resource use. This paper presents \system, a novel system designed to adaptively reconfigure robot applications to meet a predefined performance specification by leveraging \emph{runtime profiling} and \emph{automated configuration tuning}. Through experiments on multiple real robots, each running a different stack with diverse performance requirements, which could be \emph{context}-dependent, we illustrate \system's efficacy in maintaining system stability and optimizing resource allocation. Our findings highlight the promise of automatic system configuration tuning for robot deployments, including adaptation to dynamic changes.
\end{abstract}

\section{Introduction}
\begin{table*}
\caption{Common robot apps: typical resource needs, execution frequency and performance metrics}\label{tab:robot_apps_resources_needs}
\small
\centering
\begin{tabular}{llllcc}\toprule
\multicolumn{1}{c}{\multirow[c]{2}{*}{\textbf{Application}}} & \multicolumn{1}{c}{\multirow[c]{2}{*}{\textbf{Frequency}}} & \multicolumn{1}{c}{\multirow[c]{2}{*}{\textbf{Resources Used}}} & \multicolumn{1}{c}{\multirow[c]{2}{*}{\textbf{Performance Metrics}}} & \multicolumn{2}{c}{\textbf{\# of topics}} \\ \cmidrule(lr){5-6}
 & & & & \textbf{published to} & \textbf{subscribed to} \\ \midrule
LiDAR processing &~ 40 - 60 Hz &I/O, CPU &Latency, Density &12 &2 \\
Navigation &~ 20 - 40 Hz &CPU, GPU  & No collisions, smooth movements  &25 &18 \\
Localization &~ 30 - 40 Hz &CPU &Accuracy, Recovery &4 &5 \\
Web dashboard &~ 10 Hz &Network &Latency/delay, throughput &3 &12 \\
Object detection &~ 1-5 Hz &CPU, GPU &Frame rate, detection accuracy &2 &2 \\
\bottomrule
\end{tabular}
\end{table*}

Service robots are increasingly being used to assist humans in everyday tasks, with commercial examples including Amazon Astro, EEVE Willow, and Misty the Robot. Currently, these robots rely on vendor-programmed functions, but we believe that future extensibility through third-party applications (apps) will unlock diverse capabilities - much like app stores (\eg Google Play Store) did for smartphones. Furthermore, the set of actively running apps can change dynamically—whether through user-driven actions like starting or stopping functionality, installation/removal of apps, or robot personalization per user~\cite{personalize-han2024llm, synapse}, thereby significantly expanding the range of workloads that can be supported on a given robot.

Real-time responsiveness is crucial for robot apps, as delays or outdated data can hamper effective functionality in dynamic environments (\S\ref{sec:default_config_bad}).  Ensuring that numerous concurrently running and dynamically changing sets of apps can effectively share the computational resources on robots while meeting their real-time goals is thus an important resource allocation problem. Most consumer-grade robots are inherently resource-constrained, meaning they cannot simply scale resources on demand—unlike cloud servers—to meet apps' performance needs. This makes the resource allocation problem in robots particularly challenging.

Robot apps' distinct attributes further add to the challenge (\S\ref{sec:bg_ros_apps}). First, robot apps have highly diverse resource requirements, performance targets, and service level objectives (SLOs). Second, robot apps operate in various environments—such as homes, workplaces, and outdoor settings—where resource demands vary significantly due to environmental variations. Third, robots frequently adapt their algorithms based on the environment~\cite{appld}, adding to resource demand volatility.

Modern operating systems (OSes)--particularly the Linux variants widely used in robotics--fail to account for these {\em context-specific} app resource needs due to reliance on simple resource allocation heuristics. A potential workaround is to over-provision resources and manually set static allocations for each app such that it performs adequately across \textit{all} contexts; however, %
this is impractical on resource-constrained systems. %

To address these challenges, we introduce \system, a novel resource allocation framework for service robots that \textit{automatically tailors} app resource allocations to dynamic contexts by building on three unique {\em opportunities} that derive from properties of robot apps, as described next. %

First, robot apps involve persistent, long-running tasks like object tracking and sensor processing~\cite{codebotler}. Thus, \system optimizes resource allocation infrequently targeting stable environments with consistent workloads, and relies on lightweight runtime monitoring to detect significant environmental or workload changes. Further, \system maintains a library of proven configurations linked to specific environments and quickly 
 reapplies them in familiar contexts, ensuring efficiency with minimal overhead.

Second, robot functionality can be divided into {\em core services} (\eg localization, navigation) and {\em non-core apps}. Building on this, \system formulates resource allocation as an optimization problem, treating core service performance as \textit{constraints} and non-core app performance as \textit{objectives}. This reduces decision variables by limiting essential service tuning, and simplifying optimization.

Third, many robot apps operate in a data-driven manner, performing computations only when new data arrives. Coarse resource limits (\eg setting max CPU usage limits) can trigger adverse effects (\S\ref{sec:adaptor_motiv}) that degrade the performance of the throttled process needlessly. \system controls data flows -- particularly for non-core apps -- via the novel {\em adaptor} abstraction (\S\ref{sec:adaptor_definition}), gracefully reducing computational demand under resource constraints without disrupting critical functions. %

\system uses Bayesian Optimization to tune the values of OS-based system-level (\cgroup) settings, as well as the thresholds for the proposed ROS-layer adaptors 
to meet a developer-defined \textit{specification} of desired robot behavior. Once deployed, \system \textit{learns} a configuration for the current workload + context that meets the specification and applies this identified configuration to the system --  without requiring code changes to the robot apps/services. We also discuss how \system monitors for context shifts that could invalidate prior configurations and re-optimizes for new workloads and environments automatically.

We evaluate \system on state-of-the-art robot platforms, and illustrate how configurations our system identified allowed a navigation and obstacle-avoidance system on the Boston Dynamics \texttt{Spot} to run on $4\times$ fewer CPUs while still meeting developer-specified performance requirements (\S\ref{sec:results-low-resources}). We then show that our solution generalizes to multiple sets of robot apps (\S\ref{sec:results-comparision-without-adaptors}), varying environments (\S\ref{sec:results-env-change}) and other robots (\S\ref{sec:results-other-robots}).  %

\section{Related Work}
\noindent \textbf{Automated config search.} Automated config search has been successful at making web servers~\cite{cherrypicknsdi17,config-snob-atc24}, databases~\cite{db-config-tuning}, and cloud apps~\cite{oppertune-nsdi24} more efficient. They use bayesian optimization~\cite{cherrypicknsdi17}, bandit algorithms~\cite{oppertune-nsdi24}, causal inference~\cite{config-snob-atc24} or combinations of these techniques~\cite{selftune-NSDI23} to \textit{tune} config knobs. Cherrypick~\cite{cherrypicknsdi17}, for instance, uses bayesian optimization to efficiently identify the optimal cloud configuration (\ie, VM type, number of CPUs, etc) for big data analytics jobs. To the best of our knowledge, robot OS tuning via automated config search has not been attempted so far.

\vspace{0.1cm}\noindent \textbf{Context-dependent navigation.} %
Most work in this area primarily focuses on getting a set of "good" parameters under different environments. APPLD~\cite{appld} uses a supervised dataset of predefined contexts (\ie, different environments) and leverages it to train a supervised classifier to choose the current context which helps it to select the context-conditioned behavior cloning policy. Another work~\cite{contextdep_sota2} tries to dynamically find trajectory optimization weights for a Dynamic Window Approach (DWA) planner for straight-line and U-turn contexts. However, none of the aforementioned works can detect workload changes automatically as they do not profile the system, and so they need a way to classify the contexts a priori.

\section{Background}
In this section, we describe modern robot hardware (\S\ref{sec:bg_robot}) and key attributes of robot apps  (\S\ref{sec:bg_ros_apps}), setting the stage for the resource management challenges discussed subsequently.

\label{sec:bg}
\subsection{Robot Hardware \& Compute}\label{sec:bg_robot}

Service robots vary in form (e.g., wheeled, quadruped, humanoid) and feature diverse hardware configurations suited to their environments. Most rely on resource-constrained embedded systems, making efficient resource allocation critical for performance across workloads.
Typically, robots are equipped with (1) LiDAR or depth cameras for 3D perception and (2) an inertial measurement unit for odometry and orientation, both essential for navigation and obstacle avoidance. Some also have monocular or stereo RGB cameras. Additional hardware may be used depending on the application, such as tactile sensors for terrain-aware navigation~\cite{sterling-2023} or network cards for server communication and remote processing~\cite{robofleet-client,codebotler}.

Our framework's design and evaluation target multiple robot platforms, further detailed in \S\ref{sec:implementation}.

\subsection{Characteristics of Robot Applications} 
\label{sec:bg_ros_apps}
\noindent

Robot apps are typically built using the ROS framework, in which functionality is implemented in multiple \textit{nodes}, each of which is a separate Linux process. Nodes communicate via \textit{topics}, a message-passing system that enables publishing and subscribing to named queues. Robots run a ``\textit{stack}"—a cohesive set of ROS apps—continuously to perform their tasks.

\begin{figure}[t!]
    \centering
    \includegraphics[width=0.99\linewidth]{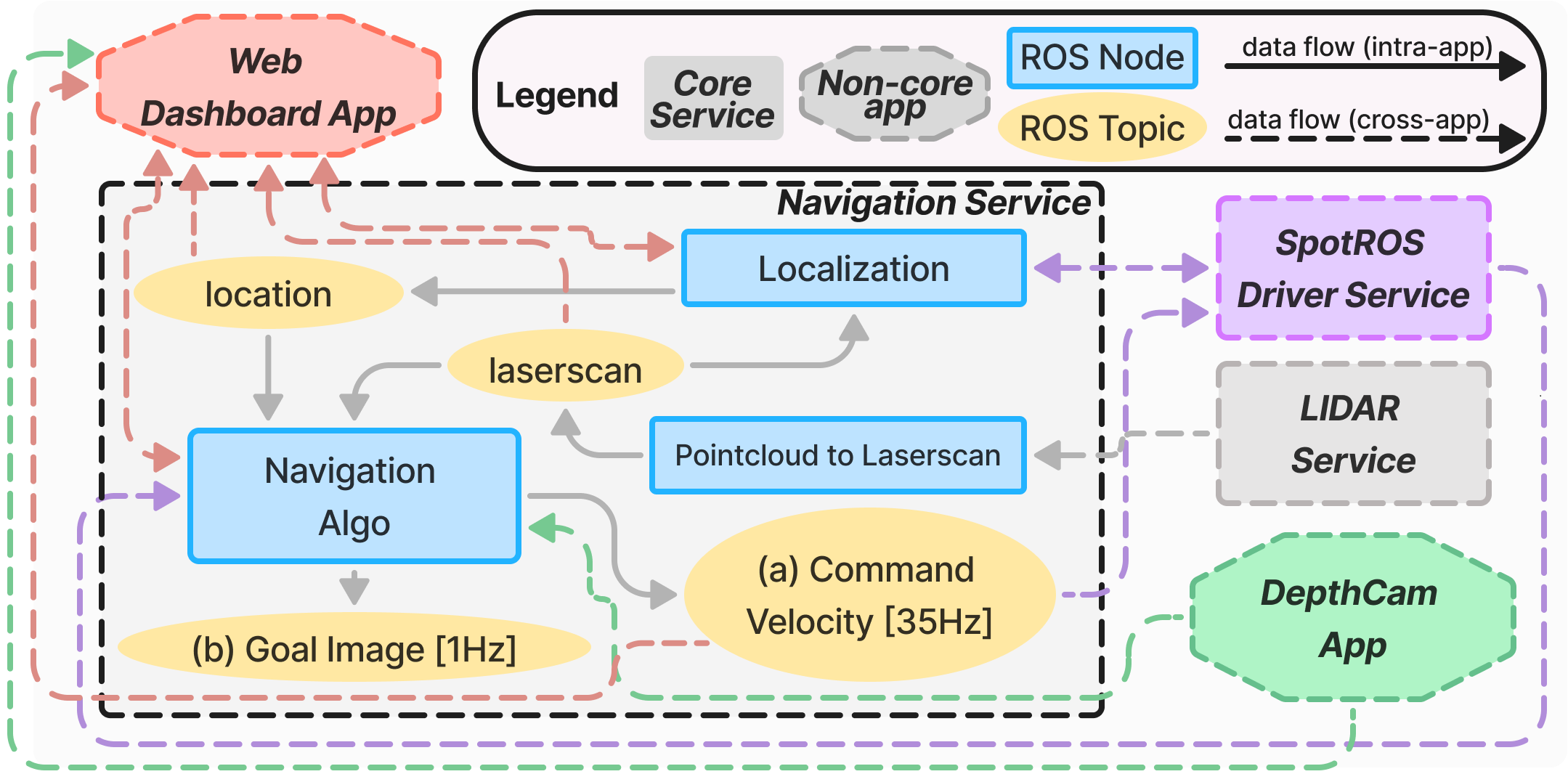}
    \caption{\robotStack{Basic-NAV} stack. A basic \textit{stack} (\ie, set of robot apps) w/ sensor processing, navigation, and telemetry~\cite{robofleet-client} via a web dashboard. Navigation is a \textit{core} service and produces two outputs: (a) \textbf{command velocity} for actuators and (b) a \textbf{goal image} showing the projected path for monitoring and debugging. The minimum safe update rate required for (a) is 35Hz (developer-specified); higher refresh-rates are possible if resources permit. All communication shown in the Figure (arrows) happens via ROS topics; we have omitted topic names for cross-app data flows (dashed arrows) for brevity.%
    }
    \label{fig:nav-app-highlevel}
\end{figure}

Fig.~\ref{fig:nav-app-highlevel} illustrates the structure of a \robotStack{Basic-NAV} stack, which we use as a running example. This stack consists of five apps, communicating through publisher-consumer dependencies (dotted lines). While simplified in the figure, each node is highly complex.  For example, the “navigation algorithm” node—one part of the navigation app—publishes to 21 topics, subscribes to 16, and employs a worker thread pool for concurrent computation.  

Beyond this complexity, robot apps have unique characteristics that complicate resource allocation:

\begin{itemize}
\item \textbf{Diverse resource demands and execution needs:} As shown in Table~\ref{tab:robot_apps_resources_needs}, robot apps vary widely in resource usage, execution rates, and performance requirements. Some rely mainly on CPUs (\eg, localization), others on GPUs (\eg, object detection), and some on both (\eg, navigation). Additional resources like networking (\eg, web dashboards) or I/O (\eg, LiDAR) may also be critical. Execution rates differ significantly: navigation runs at $20$–$40$Hz, while object detection operates $4$–$40\times$ slower. \textit{Unlike traditional workloads (\eg, databases, cloud computing), robot applications exhibit extreme resource heterogeneity, often coexisting on the same machine.} Cloud schedulers typically consolidate homogeneous workloads~\cite{tetris,resourcecentral} onto the same machine.

    \item \textbf{Implicit environment dependencies:} A robot's physical environment greatly influences resource consumption. Tasks like pose estimation~\cite{pose-estimation-topdown-1,pose-estimation-topdown-2} and object tracking~\cite{multi-object-tracking} involve multi-stage pipelines—first detecting objects, then estimating poses—leading to higher compute costs in dynamic or crowded settings. Similarly, telemetry systems~\cite{compressed-telemetry} compress static scenes efficiently but generate larger, compute-intensive streams in dynamic environments~\cite{edge-cloud-robotics,fogros2}. Thus, \textit{static resource allocations are ineffective; efficient management requires continuous adaptation to environmental changes.}  

\item \textbf{Environment-adaptive algorithm selection:} Robot developers often choose algorithms based on environmental context, further amplifying resource variability. For instance, social navigation~\cite{socialnavsurvey} may be necessary in crowded indoor spaces, while terrain-aware algorithms~\cite{sterling-2023} are better suited for outdoor environments. Algorithm selection can also occur dynamically at runtime~\cite{dynamic-algo-selection}.  Table~\ref{tab:navigation-microbenchmark} highlights how resource usage for the ``same'' navigation task varies based on algorithm. \textit{Resource allocation must not only be workload-specific but also dynamically adapt to environmental shifts.}

\end{itemize}

\noindent

\begin{table}[t!]
\centering
\caption{Resource usage of NAV algorithms (\S\ref{sec:implementation}).}
\label{tab:navigation-microbenchmark}
\small
\begin{tabular}{lll}\toprule
\multicolumn{1}{c}{Algorithm}& \multicolumn{1}{c}{Compute}& \multicolumn{1}{c}{Memory} \\\midrule
\robotStack{Basic-NAV} (Fig~\ref{fig:nav-app-highlevel}) &1.1 CPUs& 526 MB \\
\robotStack{Intermed-NAV}  &1.5 CPUs & 490 MB \\
\robotStack{Terrain-NAV} &6.9 CPUs, GPU  & 1267 MB \\
\bottomrule
\end{tabular}
\end{table}

\section{Resource allocation in modern robots}

Allocating a robot's compute resources to applications is handled by the underlying \textit{operating system} (\ie, Linux in ROS-based robots), which has multiple \textit{subsystems} (CPU, I/O, network, etc.) that coordinate resource-sharing via heuristics (\eg, the "completely fair scheduler" or CFS~\cite{linux-cfs} %
for CPU scheduling). Unfortunately, these default Linux resource schedulers often lead to degraded robot performance (\S\ref{sec:default_config_bad}) since they prioritize broad objectives (\eg, fairness), which conflict with the complex, interdependent and environment-dependent nature of robot stacks and their execution (\S\ref{sec:bg}). Linux allows configurability of these schedulers through Linux \cgroups, which enable fine-grained CPU allocation through resource limits and priorities at a per-process level. While \cgroups can improve performance by tailoring resource allocation (\S\ref{sec:cgroups_can_help}), we find that they alone cannot fully address all performance issues in robot apps (\S\ref{sec:adaptor_motiv}). We present and argue for the abstraction of configurable \textit{adaptors} (\S\ref{sec:adaptor_definition}), which modulate data flows between ROS nodes serving as important knobs for performance and stability. 

\subsection{Default OS configurations frequently lead to performance degradation in resource-constrained robots}
\label{sec:default_config_bad}

We study the performance offered by the default OS configuration when running (i) \robotStack{Basic-NAV} (Fig~\ref{fig:nav-app-highlevel}) and (ii) \robotStack{Basic-NAV} along with an object detection (\textsf{obj}) application~\cite{grounded-sam-2024} on the Boston Dynamics \texttt{Spot} robot. %

\begin{figure}
    \centering
    \begin{subfigure}[b]{0.5\linewidth}
        \centering
        \includegraphics[width=\linewidth]{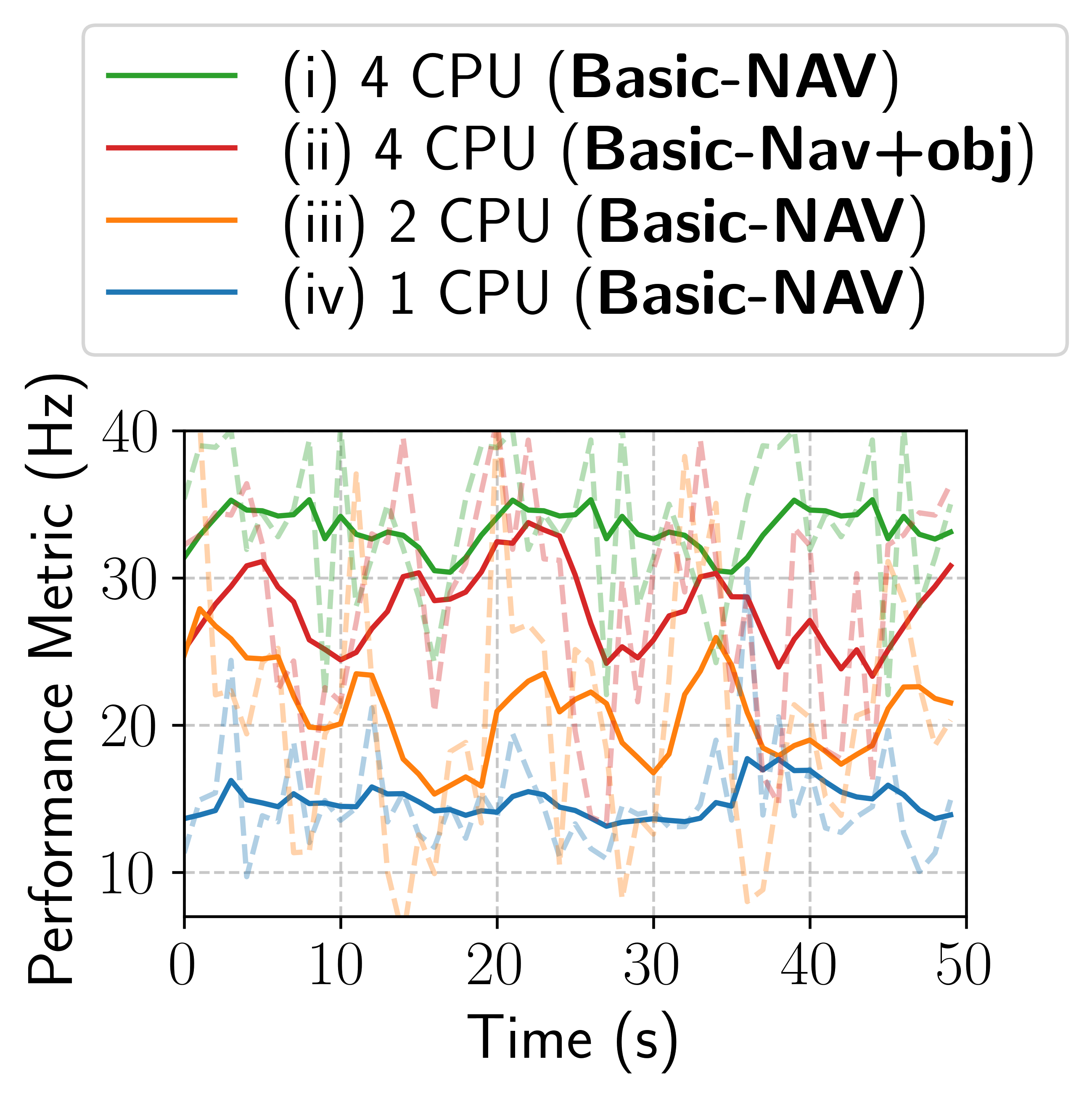}
        \caption{Update rate of command velocity in resource-constrained deployments of \texttt{Spot}. None of these deployments (ranging from 1--4 CPUs) satisfy the requirement of $>35$Hz consistently; as a result, \texttt{Spot} experiences performance degradation (Table~\ref{tab:types_of_bad_behav}).}
        \label{fig:default_bad_behav}
    \end{subfigure}
    \hfill
    \begin{subfigure}[b]{0.48\linewidth}
        \centering
        \includegraphics[width=\linewidth]{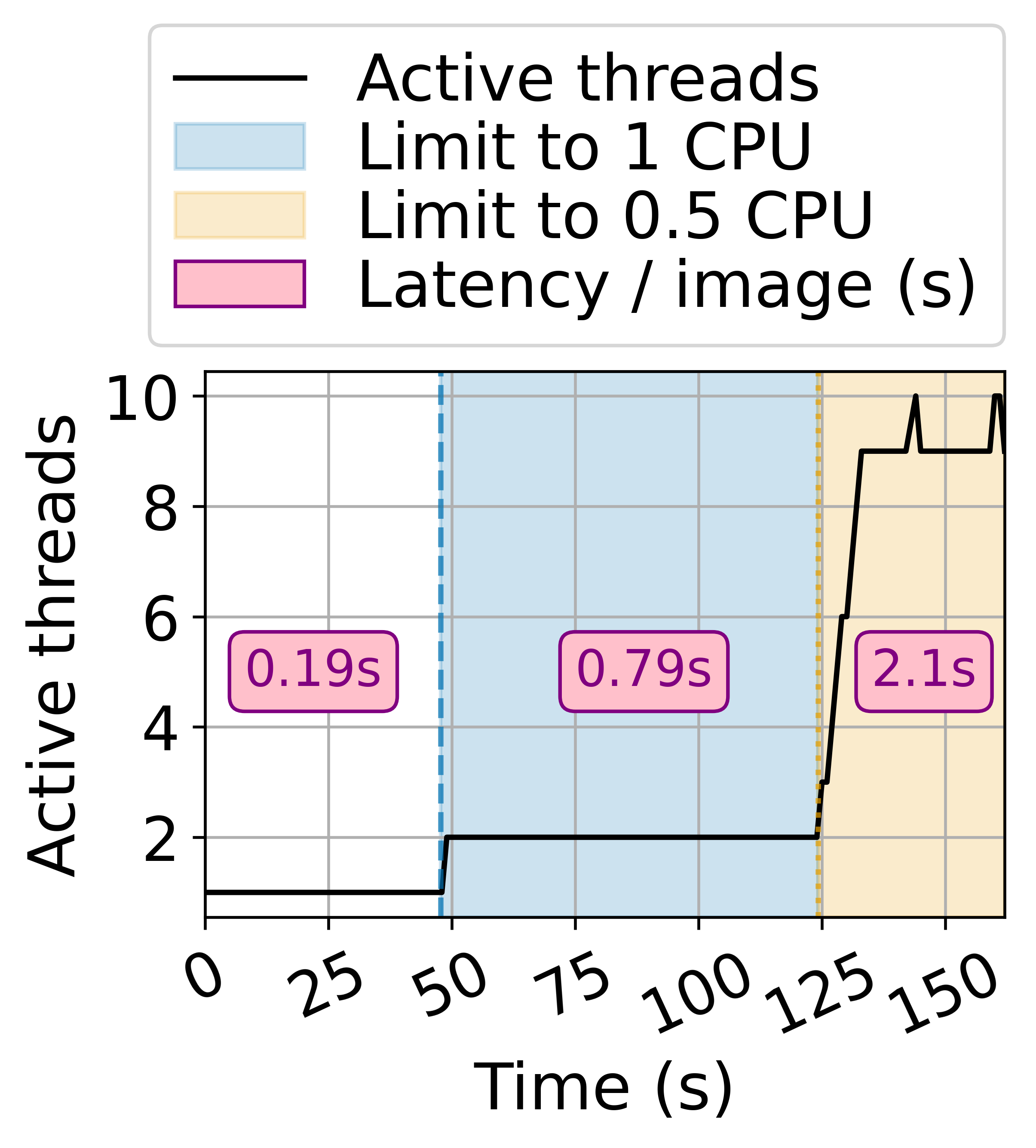}
        \caption{\robotStack{Basic-Nav}+\textsf{obj} stack on \texttt{Spot}. Number of active threads in the object detection~\cite{grounded-sam-2024} app under increasingly constrained CPU allocation. At 0.5CPUs, latency exceeds 2s as threads compete for a drastically reduced CPU budget.}
        \label{fig:queue_length}
    \end{subfigure}
    \caption{Impact of resource constraints}
    \label{fig:motivation_figure}
\end{figure}

Figure~\ref{fig:default_bad_behav} shows \textit{how frequently} new navigation decisions are sent to the actuators in settings with varying resource constraints. In a resource-rich system, without any memory or compute limits, \robotStack{Basic-NAV} was designed to update these \textit{command velocities} at 35-40 Hz, a condition that is even close to being satisfied only when $>4$ CPUs are available, as shown in Fig~\ref{fig:default_bad_behav}-i. When we constrain the system to fewer CPUs (Figs~\ref{fig:default_bad_behav}-iii,~\ref{fig:default_bad_behav}-iv) or introduce an additional app (Fig~\ref{fig:default_bad_behav}-ii), we observe that the default OS configurations cause the frequency of updates to fall by up to a factor of $3\times$. In particular, the reduced frequency of execution of the core app leads to degraded robot performance, resulting in end-to-end failures summarized in Table~\ref{tab:types_of_bad_behav}.
Here,  the default OS configurations do a poor job of allocating {\em resources that are available} across running jobs, inducing destructive inter-app contention which we show below is avoidable. %

\begin{table}\centering
\caption{Performance issues in resource-constrained robots}\label{tab:types_of_bad_behav}
\small
\begin{tabular}{p{1.6cm}p{5.5cm}}\toprule
\multicolumn{1}{c}{Robot}&Observed degradation in functionality \\\midrule
\circled{1} \texttt{Spot} &Stumbling, inability to stabilize at destination, intermittent pauses during navigation.  \\
\circled{2} \texttt{Jackal} &Collisions with obstacles \\
\circled{3} \texttt{Cobot} & Delays, jerkiness in manipulating objects.  \\
\bottomrule
\end{tabular}
\end{table}

\subsection{\cgroup settings can help core app performance}
\label{sec:cgroups_can_help}
Control groups (\cgroups) are a core feature of Linux enabling fine-grained resource management. They allow administrators to define CPU, memory, and I/O usage limits and priorities for processes, which the kernel’s scheduling mechanisms enforce at runtime. The \texttt{cpu.max} controller, in particular, allows setting the maximum fraction of CPU time that can be allotted to a process over a time window. \cgroups can form a useful mechanism to mitigate application contention in the previous section. %

For instance, the performance degradation seen in Fig~\ref{fig:default_bad_behav}-(ii) could trivially be solved by starving the object detection app. In practice, robot developers today manually identify an appropriate \verb|cpu.max| limit for the new app, ensuring it does not overuse resources, through a combination of trial and error and intuition.
As workloads and environments change, these static allocations require frequent reconfiguration, which is challenging once the robot has been deployed to run in the real-world.  We will see in \S\ref{sec:results-comparision-without-adaptors} that \system can automatically set \cgroup settings to improve performance only up to a certain extent.

\subsection{\cgroups on their own are not sufficient}
\label{sec:adaptor_motiv}
\noindent 

Unfortunately, as we show next,  relying solely on CPU throttling via \cgroups leads to suboptimal system behavior.

\vspace{0.1cm}\noindent \textbf{Internal contention within apps.} To illustrate the limitations of \cgroups, consider the \robotStack{Basic-Nav} stack (Fig~\ref{fig:nav-app-highlevel}). Some parts of the navigation app, which process LiDAR data and compute the navigation command to send down to actuators (\textbf{Task 1}), are clearly \textit{core} parts of the stack. However, the same app also overlays the planned path on an image from the camera (``Goal Image'' in Fig.~\ref{fig:nav-app-highlevel}) for monitoring, logging and debugging (\textbf{Task 2}) - useful but non-core functionality. In resource-constrained environments, \textbf{Task 2} can degrade \textbf{Task 1} by competing for the same compute resources.  %

\vspace{0.1cm}\noindent \textbf{Shared worker pools.} In the above example, we would ideally deprioritize threads handling \textbf{Task 2} to ensure \textbf{Task 1} remains unaffected. However, this is challenging because both tasks share a single thread pool: the same thread may serve either task, depending on when data arrives. This scheduling behavior is typical in C++ ROS applications using \texttt{ros::AsyncSpinner} (or \texttt{rclcpp::executors::MultiThreadedExecutor} in ROS2), which dispatches callbacks across a set of threads. A similar issue arises in Python nodes relying on \texttt{rospy.spin()}, commonly used to implement multithreaded callbacks. Both APIs serve as the \textit{de facto} approach for implementing the perpetual event loop in ROS, making them ubiquitous in robot apps. 
 
\vspace{0.1cm}\noindent \textbf{Oversubscription under throttling.}
A \cgroup-based controller for the \robotStack{Basic-Nav}+\textsf{obj} stack might reduce CPU shares for the object detection (a non-core) app whenever a core app is at risk. Object detection is a multithreaded app that uses a thread-pool of workers to process images; restricting the CPU usage of a multithreaded app can trigger a surge in parallel tasks, causing severe performance drops. As shown in Figure~\ref{fig:queue_length}, lowering the object detection app’s CPUs from 4 to 1 and then 0.5 drastically increases the number of active threads (capped at 10), raising image-processing latency beyond two seconds per frame. This occurs because each incoming image spawns a new thread. Under tighter CPU constraints, multiple threads remain active simultaneously, contending for an ever-diminishing resource pool and compounding the performance penalty. 

In both the examples in this section, direct control over data flows could be used to diminish these performance penalties. In the shared worker pool case, for instance, we could throttle the number of images being sent to \textbf{Task 2} -- the lesser images received by \textbf{Task 2}, the lesser its' compute footprint. Similarly, in the oversubscription example, throttling the number of incoming images could help prevent the compounding penalties. \textit{Adaptors}, described in the next section, are an abstraction we built to modulate data flows such as these in a \textit{transparent} way (\ie not requiring application code changes).

\section{Throttling data flows to control resource usage}
\label{sec:adaptor_definition}

\begin{figure}[t]
    \centering
    \includegraphics[width=0.84\linewidth]{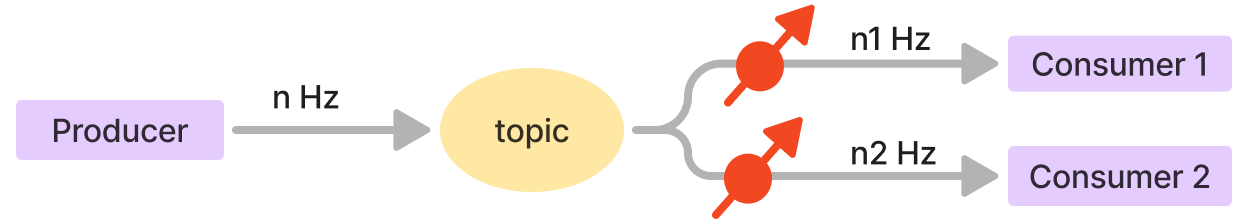}
    \caption{In traditional ROS message passing, every consumer receives all messages published to the topic (\ie $n=n_1=n_2$). Our custom ROS build places \textit{adaptors} (\adaptor) on each subscription edge, enabling independent rate control per-consumer.}
    \label{fig:adaptor_impl}
\end{figure}

\textit{Adaptors} (Fig.~\ref{fig:adaptor_impl}) offer a finer-grained resource control mechanism. In the above example scenario, adaptors can ``intercept'' camera data on a per-subscriber basis, throttling (\ie dropping data) only the problematic paths without penalizing everyone else. Adaptors thus enable ROS to manage data flows within and across apps and achieve more nuanced resource control than \cgroups alone can provide. Adaptors are inserted on each subscription edge and they control the frequency and volume of messages passed to each consumer. Both attributes can be dynamically reconfigured at runtime to shape the resource usage pattern of the consumer. We integrate adaptors into the \verb|ros_comm| library, a core ROS component responsible for communication, by modifying the subscriber API to enable message filtering. Adaptors are automatically setup when an app running on our modified ROS build calls \verb|rospy.Subscriber| (Python) or \verb|nh.subscribe| (C++). Each adaptor also has a dedicated ROS service associated with it, allowing dynamic adjustment of its filtering frequency at runtime. Our implementation required only $\sim150$ lines of code across the C++ and Python implementations of \verb|ros_comm|.

The problems we described in \S\ref{sec:adaptor_motiv} can be fixed by changing the application code itself; for example, a developer could create separate worker pools for each independent task in the node, or split out the node into multiple nodes. However, these approaches are intrusive, forcing developers to re-architect their apps. Adaptors, by contrast, offer an  application-transparent, drop-in solution. We will show in \S\ref{sec:results-low-resources} and \S\ref{sec:results-comparision-without-adaptors} that the addition of adaptors allows \system to unlock higher performance over a \system-like approach that uses only \cgroups.

\section{\system Design}\label{sec:approach}
The previous section demonstrated how resource constraints degrade robot stacks (§\ref{sec:default_config_bad}), and how carefully tuning CPU limits (§\ref{sec:cgroups_can_help}) or inserting adaptors (§\ref{sec:adaptor_definition}) can restore performance. However, deciding precisely which \cgroup parameters or adaptor rates to apply remains challenging, often requiring expert insight into application dependencies and resource demands and painstaking and error-prone manual tuning. Even after identifying the best configuration, we will show (\S\ref{sec:results-config-reuse-bad}) that almost any change in the workload renders the existing configuration suboptimal for the new scenario. \system, which can automatically pinpoint ``good'' configurations for complex app mixtures running on a robot at any given time and environment is proposed as a solution to this challenge.

\subsection{Problem Definition}\label{sec:problem_defn}
Consider a robot stack (\ie, the set of applications $a_1$, $a_2$, $\ldots$, $a_n$) operating under specified conditions, which include both system resources and the physical environment. For each application, we assume the developer provides a performance metric ($\mathbf{J}_i$) to evaluate its behavior (\eg, frame rate, message publish frequency) and a \textit{desirable target} value $J_i^o$. Achieving this target ensures the robot operates effectively. This assumption is reasonable, as such metrics are often easy to define.

As mentioned earlier, we can 
categorize robot apps into two classes: 
\begin{itemize}
    \item {\bf Core services ($A_c$):} Critical for the robot’s functionality (\eg, localization, navigation, obstacle avoidance) and must always perform reliably.
    \item {\bf Non-core apps ($A_n$):} Non-essential and can be executed opportunistically when resources permit.
\end{itemize}
Armed with this categorization, we can specify the task of identifying an optimal configuration as a constrained black-box multi-objective optimization problem:
\begin{align*}
    \argmax_c &\min (\mathbf{J}_i(c), J_i^o) \,\,\,\,\,\,\,\, \forall \, a_i \in A_n \\
    &\text{s.t.}\,\,\,\,\,\, \mathbf{J}_i(c) \geq J_i^o \,\,\,\, \forall \, a_i \in A_c \\
    &c \in \mathbf{C} \,\,\,\, \text{\ie, config knobs space}
\end{align*}
where $c$ is one specific \textit{configuration} of the operating system (\ie values for \cgroups) and 
 for adaptors from the entire possible configuration space ($\mathbf{C}$).

The above defines an optimization problem over $\mathbf{C}$ with $|A_n|$ objectives and $|A_c|$ constraints, with the full solution existing on a Pareto front. The $\textsf{min}$ operator enforces an upper bound on performance for each application, ensuring that once an app reaches its target, additional resources are not wasted on it. For apps without a strict target, this bound is effectively set to $\infty$. When no configuration that satisfies the constraints for the core apps ($A_c$), the optimization process outputs \textsc{UNSAT}, indicating none of the explored configurations are feasible, meaning that the robot developer has to either relax their constraints or provision more resource.

\subsection{The \system Approach}

\begin{figure}[t!]
    \centering
    \includegraphics[width=\linewidth]{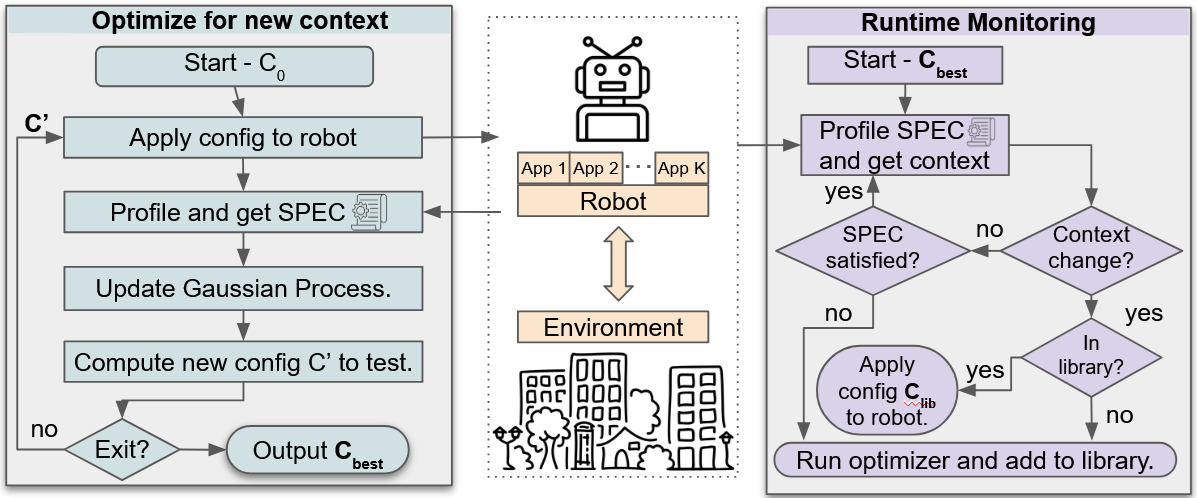}
    \caption{\system flowchart: How \system learns configs (given a specification -- \textsf{SPEC}) for new contexts and performs runtime monitoring to identify when relearning is needed.}
    \label{fig:overall}
\end{figure}

\noindent Figure~\ref{fig:overall} outlines our framework with details provided next. %
\vspace{0.1cm}

\noindent
{\bf Specification and configuration space.} The robot developer begins by defining performance metrics ($\mathbf{J}(c)$) for their apps, along with desirable target values ($J^o$) for each. For example, the developer might say that the navigation should run at a frequency (\ie, $\mathbf{J}(c)$) of more than 35 Hz (\ie, $J^o$). These specifications form the basis of the optimization problem. Our system then populates a list of available config knobs -- we place each ROS node in a separate \cgroup, allowing our system to control the \verb|cpu.max| allocations for nodes independently. Additionally, we place adaptors on topics that (A) have high volumes of messages (\ie $> 5$ Hz), \textbf{and} (B) that involve a non-core app either as a publisher or subscriber. We do this for for practical reasons (\ie cut down on the number of config knobs to tune); without these conditions (A) and (B) in place even the \robotStack{Basic-NAV} stack (Fig~\ref{fig:nav-app-highlevel}) would have $50+$ adaptors making optimization unwieldy. This design choice is justified, as the primary goal of adaptors is to limit excessive resource usage by large data flows involving non-core apps.

\vspace{0.1cm}\noindent
{\bf Initial learning.} Using the provided specifications, \system models the requirements of core services as constraints and those of non-core apps as optimization objectives. As shown in the left half of Fig~\ref{fig:overall}, \system leverages Bayesian optimization, well-suited for efficiently exploring configurations in high-dimensional, black-box scenarios. The optimization algorithm samples the config space ($\mathbf{C}$) to identify a config $(C)$ that is applied to the system; we then profile the performance of our objectives and constraints under $C$ for a brief amount of time ($5$s), after which we pass these observations to the optimizer; the optimizer uses this information to suggest a new point $C'$. This iterative process outputs the best config identified $C_{best}$ (or \textsf{UNSAT} if no solutions are found).

\vspace{0.1cm}\noindent
\textbf{Online monitoring and relearning.} The right half of Fig.~\ref{fig:overall} illustrates the online monitoring process after a suitable configuration is identified and applied. \system uses a three-layer monitoring system including: (a) an eBPF~\cite{ebpf} monitor that runs continuously, (b) a \verb|rosmaster|-based monitor that wakes up every $30$ seconds, and (c) a lightweight constraint monitor that continuously checks if the developer defined-spec is being met. More details are discussed through an illustration in the results section~\ref{sec:results-env-change}. The eBPF and \verb|rosmaster|-based monitors detect new processes and ROS nodes respectively, and raise warning flags before constraint violations are detected by the constraint monitor.

\vspace{0.1cm}\noindent \textbf{Optimizations}. To minimize downtime during retraining, \system maintains a configuration \textsf{library}, mapping operational contexts (\eg current location coordinates, list of apps) to optimal configurations, enabling rapid reuse of known good configs; a design pattern that other robot systems use as well~\cite{appld}.
When immediate retraining is not feasible (\eg the robot is executing a critical task and cannot pause), we defer retraining by logging a trace of the current inputs (capturing sensor inputs via \verb|rosbag|) and applying quick remediation by terminating non-core apps and reallocating their CPU shares to core services.

\section{Implementation}\label{sec:implementation}
\noindent \textbf{Robots.} We implement \system on three different robots: (a) \textsc{Spot}: Boston Dynamics Spot w/ NVIDIA AGX Orin, (b) \textsc{Jackal}: Clearpath Jackal w/ Intel i7 8 core machine, and (c) \textsc{Cobot}: a mobile navigation base with a Kinova arm, w/ Intel NUC 10 (no GPU). We primarily use \textsc{Spot} for evaluating \system, though also show that it generalizes easily and effectively to other robots.

\vspace{0.1cm}\noindent \textbf{Stacks.} We evaluate \system on six different sets of stacks across the three robots. For \textsc{Spot}, we have (i) \robotStack{Basic}, which uses standard DWA based local planner for obstacle avoidance, and a carrot-based high level planner, (ii) \robotStack{Intermed}, which adds an additional layer of discretized A-star grid-based planning in between the two layers described in \robotStack{Basic}, and (iii) \robotStack{Terrain}, that uses the terrain-aware navigation stack~\cite{sterling-2023}. For \textsc{Cobot}, we have \robotStack{CoNav} which is similar to \robotStack{Basic} but adapted for \textsc{Cobot}, and \robotStack{CoMap}, a manipulation stack based on top of Kinova Kortex API. For \textsc{Jackal}, we use \robotStack{Phoenix} which is Army Research Laboratory's navigation stack~\cite{phoenix-arl}. In addition, we use these non-core apps: web dashboard (\textsf{web})~\cite{robofleet-client}, object detection (\textsf{obj})~\cite{grounded-sam-2024}, segmentation (\textsf{segment}) and pose estimation ($\textsf{pose}$)~\cite{yolov11}. 

\begin{figure}[t]
    \centering
    \includegraphics[width=0.5\linewidth]{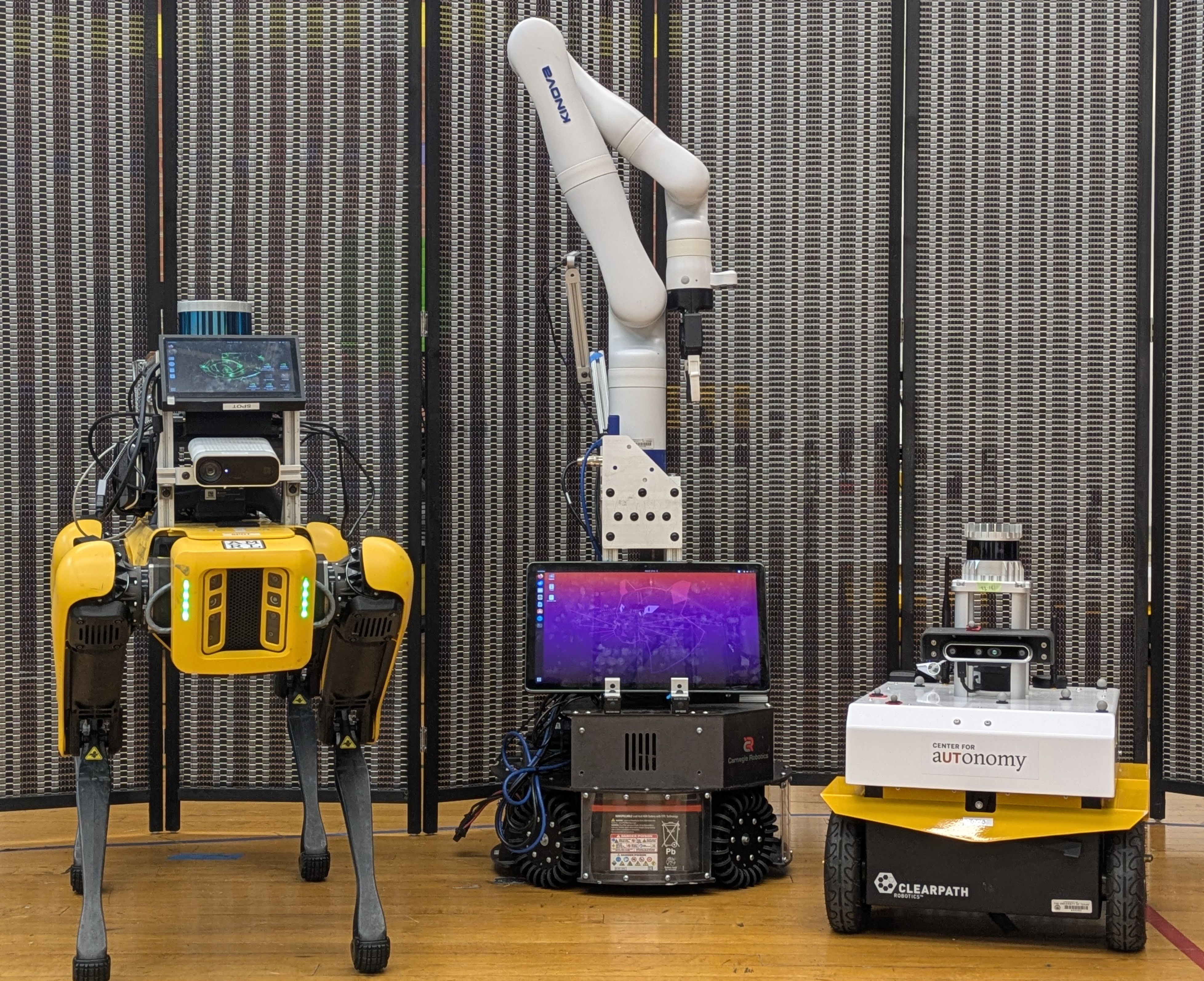}
    \caption{Robots we use: Spot, Cobot, and Jackal (left to right).}
    \label{fig:all-robots}
\end{figure}

\section{Evaluation}
\label{sec:eval}

\noindent In this section, we show: \begin{itemize}
    \item A walkthrough of \system's learning process on \robotStack{Basic}, illustrating how \system discovers configurations that outperform default OS configurations, even when the default OS is equipped with $4\times$ more resources (\S\ref{sec:results-low-resources})
    \item Results on \texttt{Spot} demonstrating \system identifies good configurations for multiple robot stacks (\S\ref{sec:results-comparision-without-adaptors})
    \item Demonstration of how the optimal configuration for one set of apps doesn't work for others (\S\ref{sec:results-config-reuse-bad})
    \item Illustration of how \system handles runtime monitoring and retraining (\S\ref{sec:results-env-change})
    \item \system's generality to other robots (\S\ref{sec:results-other-robots})
\end{itemize}

\noindent \textbf{Metrics}. We measure performance using two metrics: for core services, the \textbf{\textit{constraint satisfaction rate}} measures the proportion of time in which all constraints are satisfied -- ideally, we would want this to be 100\%; for non-core apps, we use the developer-specified \textbf{\textit{performance of non-core app}} to compare configs. For both these metrics, higher values are better, although a high non-core performance at the cost of a lower-satisfaction rate is undesirable.

\vspace{0.1cm}\noindent \textbf{Comparisons}. We compare the performance of the \system-tuned configs with (a) the default config and (b) \system-\textsf{cg}, \ie a \system without adaptors (tunes only \cgroup settings), to quantify the relative improvements caused by adaptors.

\subsection{\system: learning a config for \robotStack{Basic} on \texttt{Spot}}
\label{sec:results-low-resources}
\noindent~Table~\ref{tab:resultsA-performance-under-1-cpu} below summarizes the configurations identified by \system, the satisfaction rate of those configs, and the non-core performance. \system performs nearly $3\times$ better than the default config, while maintaining a near-perfect satisfaction rate. In fact, the default config fails to outperform the \system-tuned config on constraint satisfaction rate, \textit{even when we provide the default with 4$\times$ more CPUs}.

\begin{table}
\centering
\caption{\system on \robotStack{Basic} with \textsf{web}}
\label{tab:resultsA-performance-under-1-cpu}
\small
\begin{tabular}{lcc}\toprule
& \multicolumn{1}{c}{Core (\%CS)} & \multicolumn{1}{c}{Non-core (Hz)} \\\midrule
\system & \textbf{99.42\%} & 9.90 \\
\system-\textsf{cg} & 93.76\% & 1.97 \\
Default & ~0.00\% & 3.22 \\
\midrule
Default ($4\times$ CPUs) & ~84.94\% & 29.53 \\
\bottomrule
\end{tabular}
\end{table}

\begin{figure}
    \centering
    \includegraphics[width=0.85\linewidth]{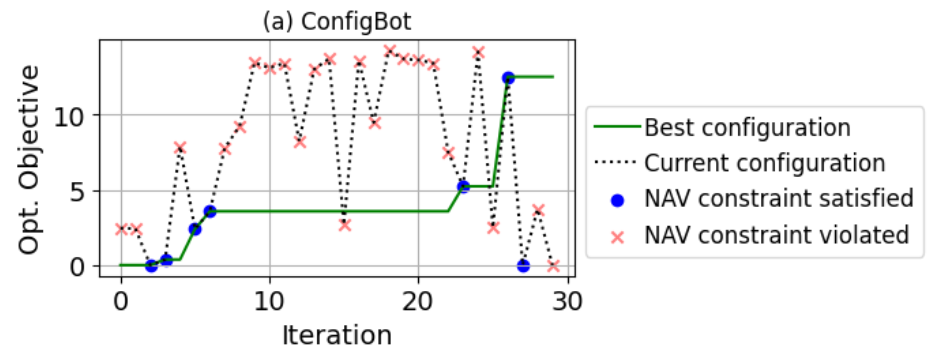}
    \caption{Optimizing \robotStack{Basic} with \system on \texttt{Spot}}
    \label{fig:resultsA-optimization-progress}
\end{figure}

Fig~\ref{fig:resultsA-optimization-progress} plots the optimization progress of \system as it identifies the config in Table~\ref{tab:resultsA-performance-under-1-cpu}. In Fig~\ref{fig:resultsA-optimization-progress}, each point corresponds to one configuration tested by the optimizer; a red cross indicates that config fails to satsify the core service constraint. The optimizer chooses the highest performing blue-dot (\ie constraints satsified) as the best config; the \textit{green} solid line tracks the best config identified \textit{so far}.

\begin{figure}
    \centering
    \includegraphics[width=0.8\linewidth]{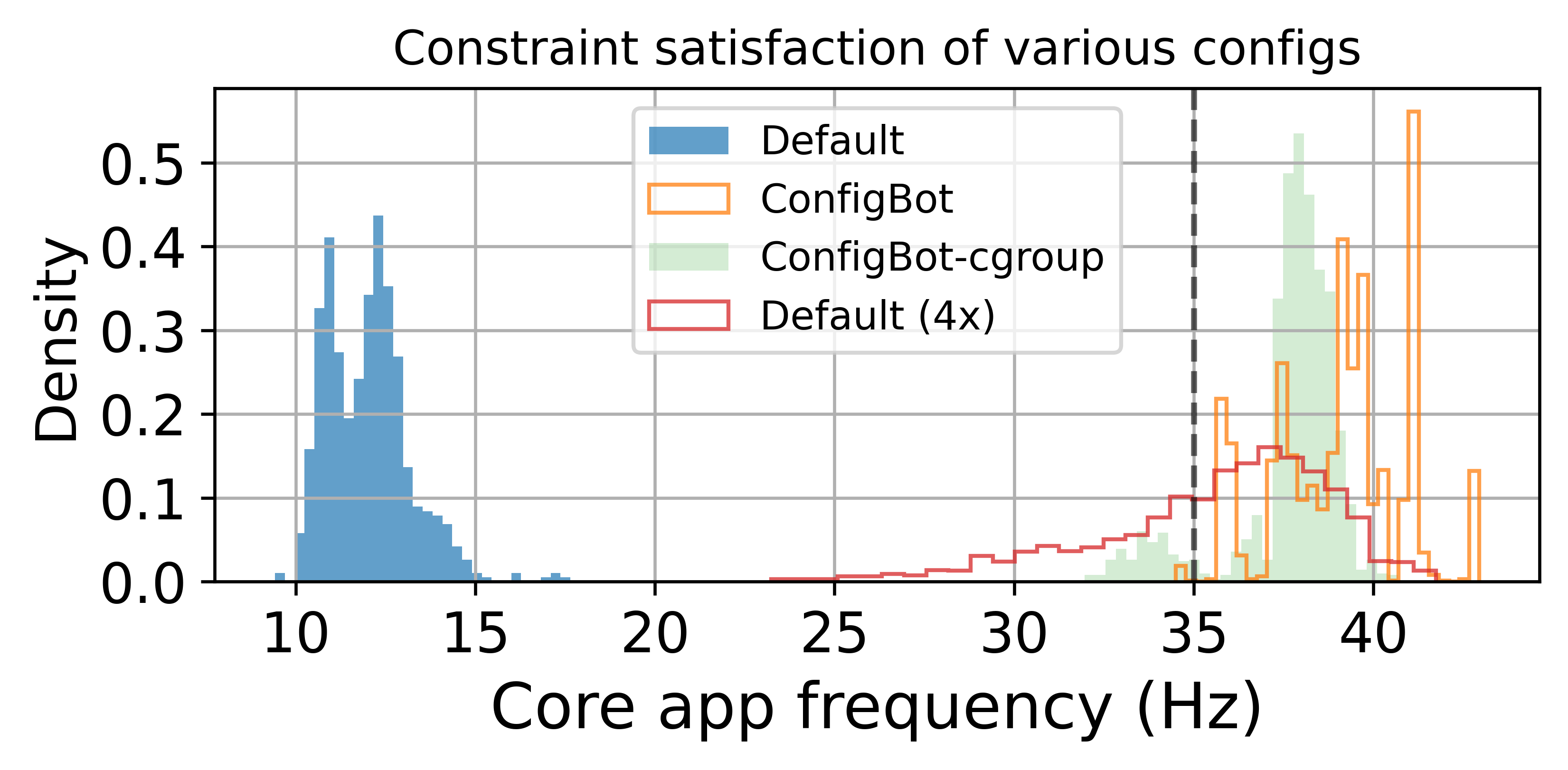}
    \caption{Histogram: performance of core-services}
    \label{fig:cs_density}
\end{figure}

The constraint satisfaction rate is a binary metric - which tells us if a config satisfies the constraint, or doesn't; it doesn't inform us on how \textit{close} it was to satisfying the constraint threshold. 
Fig~\ref{fig:cs_density} plots a histogram of the core services' performance and shows that the default config ($10-15$Hz) is  far from the developer-defined performance constraint ($35$Hz).

\subsection{Benchmarking \system performance on \texttt{Spot}} \label{sec:results-comparision-without-adaptors}

Table~\ref{tab:results_random_default_configbot} summarizes the performance of \system in optimizing three different navigation stacks on \texttt{Spot}, while an additional resource-hungry non-core app (\texttt{Obj}) is running. The number in (parentheses) represents the performance of the non-core app (\texttt{obj}) and \%ages are satisfaction rates.

\begin{table}
\caption{Performance of \robotStack{NAV} stack with \robotStack{obj}.} %
\label{tab:results_random_default_configbot}
\small
\centering
\begin{tabular}{lccc}
\toprule
 & \robotStack{Basic} & \robotStack{Intermed} & \robotStack{Terrain} \\
\midrule
\system & \textbf{100\%} (\textbf{2.12}) & \textbf{100\%} (\textbf{1.79}) & 96.0\% (\textbf{1.35}) \\
\system-\textsf{cg} & \textbf{100\%} (1.78) & \textbf{100\%} (1.71) & \textbf{100\%} (0.70) \\
Default & 22.4\% (1.45) & 0.0\% (1.20) & 0.4\% (0.46) \\
Random-1 & 99.3\% (0.62) & 0.0\% (0.62) & 99.4\% (0.56) \\
Random-2 & 35.0\% (1.04) & 0.0\% (0.95) & 24.9\% (1.04) \\
Random-3 & 86.9\% (1.31) & 60.2\% (1.61) & 39.2\% (0.80) \\
\bottomrule
\end{tabular}
\end{table}

Table~\ref{tab:results_random_default_configbot} illustrates how \system's config has a near 100\% satisfaction rate for all three stacks while being able to perform between 49\% (for \robotStack{Intermed}) to 193\% (for \robotStack{Terrain}) better on the non-core tasks, when compared against default. Additionally, the addition of adaptors allows \system to discover configs that have as much as 92\% (\robotStack{Terrain}) better performance on non-core apps when compared against \system-\textsf{cg}. In addition to the usual comparisions, Table~\ref{tab:results_random_default_configbot} also includes three randomly chosen configurations sampled uniformly at random from the config space -- none of these configs were able to preserve high constraint satisfaction while maximizing non-core performance to the extent \system-tuned configs were able to, illustrating the sparsity of good configs  within the config space.

\subsection{Absence of a globally optimal config}
\label{sec:results-config-reuse-bad}

Throughout this paper, we have been alluding to the benefits of constantly learning and relearning configurations for each new context. To illustrate the importance of relearning, we attempt to see if there exists an optimal config that works across all three of our stacks. Intuitively, a config we learn for a \emph{more-resource intensive} core service (\eg \robotStack{Terrain}) should perhaps work well on simpler apps (\eg \robotStack{Basic}, \robotStack{Intermed}). 

Table~\ref{tab:config_reuse_is_bad} evaluates the best config we identified for each app (\eg $\robotStack{Terrain}_c$) on the other two apps (\eg \robotStack{Basic}). We find that there is a clear one-one correspondence between the highest performing config on each stack. Specifically, the optimal config learned from \robotStack{Basic} indeed does better on \robotStack{Basic} in comparison to \robotStack{Terrain}, validating the need for learning a new config whenever context changes and supporting the absence of a trivially transferable globally optimal config.

\begin{table}[h] %
\caption{Cross-evaluation of optimal configs on stacks.}
\label{tab:config_reuse_is_bad}
\small
\centering
\begin{tabular}{lccc}
\toprule
\multirow{2}{*}{Config} & \multicolumn{3}{c}{Navigation Stack (running on robot)} \\ 
\cmidrule{2-4}
 & \robotStack{Basic} & \robotStack{Intermed} & \robotStack{Terrain} \\
\midrule
$\robotStack{Basic}_c$ & \textbf{100\%} (\textbf{2.12}) & 85.7\% (1.63) & 69.2\% (1.51) \\
$\robotStack{Intermed}_c$ & \textbf{100\%} (1.86) & \textbf{100\%} (\textbf{1.79}) & 45.31 (\textbf{1.52}) \\
$\robotStack{Terrain}_c$ & 99.4\% (1.61) & 4.4\% (1.38)  & \textbf{96.0\%} (1.35) \\
\bottomrule
\end{tabular}
\end{table}

\subsection{Context-dependent relearning} \label{sec:results-env-change} 

\begin{figure}
    \centering
    \includegraphics[width=1.0\linewidth]{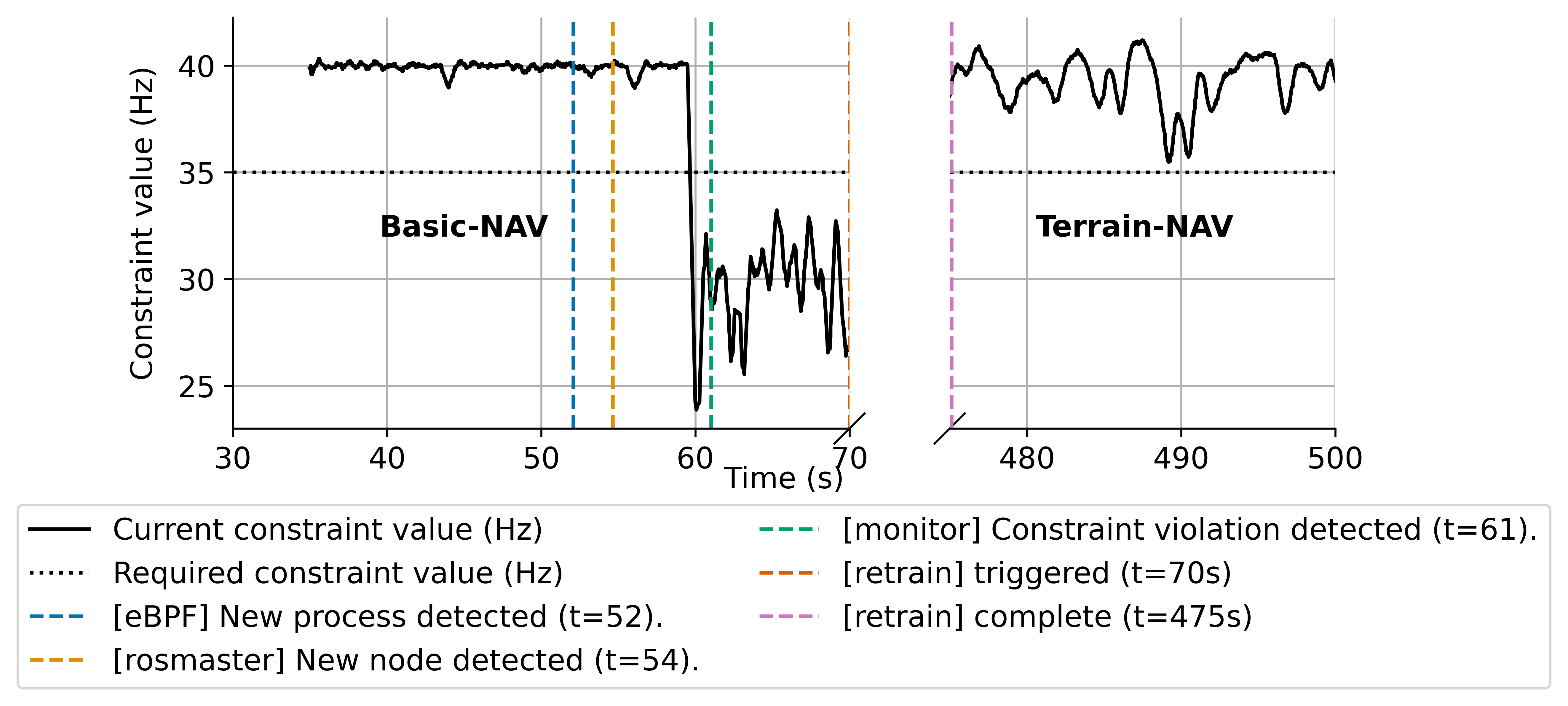}
    \caption{\system runtime monitoring + relearning}
    \label{fig:app-change}
\end{figure}

Figure~\ref{fig:app-change} demonstrates a scenario where \texttt{Spot} is initially navigating an indoor environment with \robotStack{Basic}; then, it steps outdoors and switches to \robotStack{Terrain} at time $t\approx50s$, which is what it has been programmed to do. \system is able to detect this almost instantly through the use of an eBPF (Extended Berkeley Packet Filter) monitor which enables low-overhead, in-kernel monitoring of system activity and can detect new processes even before they initialize ROS (\eg before \verb|rospy.init()|). Our eBPF monitor dispatches a \verb|wakeup| message to our second monitor, a \verb|ROS|-Python node that uses the \verb|rosmaster| API, which would otherwise check for new nodes every $\sim30$ seconds. Due to the \verb|wakeup|-call, our \verb|rosmaster|-based profiling identifies the new ROS nodes less than 2s after they are initialized; at this stage, we could either start the reoptimization process (or reuse a config from the \textsf{library}, if this context has been seen before); however, for illustrative purposes, we don't do so, and wait for constraint violations to accumulate over a period of time before our third trigger (the constraint monitor) starts the retraining. 

\subsection{Generalization to other robots} \label{sec:results-other-robots}
Table~\ref{tab:results-other-robots} summarizes the results of running a completely different set of stacks on \textsc{Jackal} and \textsc{Cobot}. \system clearly outperforms the Default config, which validates its general and widespread applicability.

\begin{table}[h]
\centering

\caption{Performance on \textsc{Cobot} and \textsc{Jackal}.}
\label{tab:results-other-robots}
\small
\begin{tabular}{llcc}
\toprule
Robot & Core (non-core) & \system & Default \\
\midrule
\textsc{Cobot} & \robotStack{CoNAV} (\textsf{pose}) & \textbf{100\%} (\textbf{0.78}) & 47.1\% (0.45) \\
\textsc{Cobot} & \robotStack{CoMap} (\textsf{pose}) & \textbf{100\%} (\textbf{2.47}) & 54.3\% (1.67) \\ 
\midrule
\textsc{Jackal} & \robotStack{Phoenix} (\textsf{web}) & \textbf{100\%} (2.32) & 7.2\% (\textbf{6.71}) \\  %
\textsc{Jackal} & \robotStack{Phoenix} (\textsf{segment}) & \textbf{98.2\%} (\textbf{0.76}) & 0.0\% (0.42) \\  %
\bottomrule
\end{tabular}
\end{table}

\section{Conclusion and Future Work}

We presented \system, an automated configuration tuning system designed to dynamically reconfigure service robots to meet predefined performance specs, and demonstrated it's ability to maintain system stability across a range of challenging scenarios. Looking ahead, we identify two key directions for future work. First, automating the process of defining performance specifications remains an open challenge. Specifically, developing methods to infer these specifications directly for robot applications from high-level task descriptions could significantly reduce the burden on developers. Second, while \system currently reacts to changes in the environment or application demands, incorporating reinforcement learning (RL) could enable proactive management of resources. By predicting potential performance bottlenecks or resource conflicts before they occur, an RL-driven controller could anticipate problems and adapt configurations preemptively.

\bibliographystyle{IEEEtran}
\bibliography{citations}

\end{document}